\documentclass[times, 10pt, twocolumn]{article} 
\usepackage{latex8}
\usepackage{times}
\usepackage{graphicx}
\usepackage{url}
\usepackage{bm}
\usepackage{dblfloatfix}    % To enable figures at the bottom of page

\begin{document}

\title{Image Captioners Sometimes Tell More Than Images They See}

\author{Honori Udo${}^{*}$\hspace{2cm}Takafumi Koshinaka\\
The School of Data Science, Yokohama City University\\
22-2 Seto, Kanazawa-ku, Yokohama 236-0027 Japan\\
 \\
%\and
{\small * Now with NTT Comware Corporation}
%First line of institution2 address\\ Second line of institution2 address\\ 
%SecondAuthor@institution2.com\\
}

\maketitle
\thispagestyle{empty}

\begin{abstract}
Image captioning, a.k.a. ``image-to-text,'' which generates descriptive text from given images, has been rapidly developing throughout the era of deep learning. To what extent is the information in the original image preserved in the descriptive text generated by an image captioner? To answer that question, we have performed experiments involving the classification of images from descriptive text alone, without referring to the images at all, and compared results with those from standard image-based classifiers. We have evaluate several image captioning models with respect to a disaster image classification task, CrisisNLP, and show that descriptive text classifiers can sometimes achieve higher accuracy than standard image-based classifiers. Further, we show that fusing an image-based classifier with a descriptive text classifier can provide improvement in accuracy.
\end{abstract}

%------------------------------------------------------------------------- 
\Section{Introduction}

Advances in neural network-based representation learning have enabled ``embedding,'' which maps any type of data into a latent space with a certain number of dimensions. Numerous methodologies have been proposed for handling media data with different modalities~\cite{Mariana,Chai}.  In particular, significant research results have recently been reported in the field dealing with images and texts referred to as ``Vision and Language'' (V\&L)~\cite{Flamingo}.  Typical tasks here would include visual question answering (VQA) for answering questions with images~\cite{VQA}, image captioning for assigning descriptions to images~\cite{Ushiku}, and image generation for generating images from descriptions~\cite{Mansimov}.

Image captioning (image-to-text), which is addressed in this paper, was inspired by sequence-to-sequence learning in neural machine translation~\cite{seq2seq} and has shown remarkable progress employing a similar approach, one in which a context vector obtained by encoding an input image is decoded to generate descriptive text in an auto-regressive manner~\cite{Vinyals}.  It has a two-sided relationship with image generation (text-to-image), which has attracted public attention with the advent of Dall-E 2 and subsequent approaches to generation~\cite{Dall-E,Imagen}.  Further development is expected with the support of such foundation models for Vision and Language as CLIP~\cite{CLIP} and BLIP~\cite{BLIP}, which are trained with a large number of images and large amount of text found on the Internet.

Image captioning has something in common with automatic speech recognition (ASR a.k.a. speech-to-text), which has long received much research attention.  While ASR converts acoustic signals caused by air vibration into text, image captioning converts light signals into text.  ASR extracts only linguistic information from input speech and discards such non-linguistic information as tone, emotion, and the gender and age of the speaker.  In this sense, ASR can be viewed as a kind of feature extraction.  Image captioning is similar in that it extracts certain information from an input image and discards other information.  We aim to clarify what that ``certain information'' actually is.

In speech emotion recognition, Srinivasan et al.~\cite{Srinivasan} employ linguistic information (text) obtained from ASR as features and show that the emotion-recognition accuracy is superior to that with conventional methods that use only acoustic features.  This may also apply in image captioning.  Could an image captioner also serve as a feature extractor that complements image recognition?  In this paper, we examine how accurately image classification can be performed using descriptive text obtained by an image captioner as a feature.  We also try to improve classification accuracy by combining it with standard image classifiers based on convolutional neural networks (CNNs).  Our experimental setup is designed for use with CrisisNLP~\cite{CrisisNLP,CrisisMMD}, a benchmark dataset for disaster image classification.  There have been earlier studies that attempted multi-task learning with image classification and captioning as two parallel tasks~\cite{Yoon}, but, to the best of our knowledge, this is the first attempt to use an image captioner as a feature extractor for image classification.

The remainder of this paper is organized as follows.  Section~\ref{sec:sysconfig} describes the configuration of our image classification system using image/text-based classifiers combined with an image captioner.  The experimental setup and results for image- and text-based single-modal systems as well as fused multi-modal systems are presented in Section~\ref{sec:experiments}.  Section~\ref{sec:conclusion} summarizes our work.

%------------------------------------------------------------------------- 
\section{System Configuration}
\label{sec:sysconfig}
\begin{figure}[tbp]
\begin{center}
\includegraphics[scale=1.0]{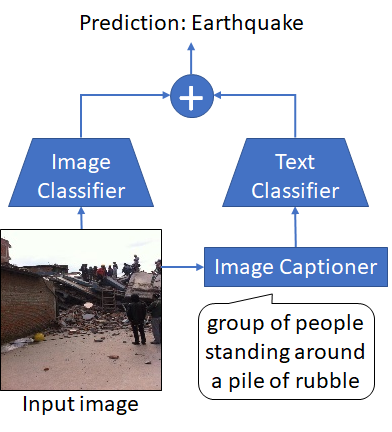}
%\includegraphics[scale=0.8, bb=0 0 200 190]{systarch2.png}
%\vspace{-5mm}
\end{center}
\caption{System configuration: a standard image-based classifier (left half) and a text-based classifier combined with an image captioner (right half).}
\label{fig:system architecture}
\end{figure}

Figure~\ref{fig:system architecture} shows the configuration of the final form of the image classification system considered in this paper.  The left half shows a standard image classifier for use with a neural network such as a CNN.  The right half shows the connection of an image captioner with a text classifier, in tandem, for the classification of images on the basis of linguistic information extracted from images.

\subsection{Image and Text Classifiers}
We used pre-trained models throughout.  MobileNetV2~\cite{MobileNetV2} and EfficientNet~\cite{EfficientNet} were used for the image-based classifier, and BERT${}_{\rm BASE}$~\cite{BERT} was used for the text-based classifier after fine-tuning with the training data for the target task.  Regarding hyper-parameter settings in fine-tuning, such as learning rate, mini-batch size, number of epochs, data augmentation, dropout, etc., we follow publicly available tutorial codes~\footnote{Transfer learning and fine-tuning \url{https://www.tensorflow.org/tutorials/images/transfer_learning}}
\footnote{Image classification via fine-tuning with EfficientNet
\url{https://keras.io/examples/vision/image_classification_efficientnet_fine_tuning/}}
\footnote{Classify text with BERT \url{https://www.tensorflow.org/text/tutorials/classify_text_with_bert}}.

\subsection{Image Captioners}
We focused on three image captioning models. None of them were fine-tuned using the data for the target task (because no image description text was available for the target task), and the original models were used as it is.
\vspace{3mm}

\noindent
{\bf InceptionV3+RNN}: This is a basic model of small scale that encodes an input image into a vector by using InceptionV3~\cite{InceptionV3} and then decodes it to generate a caption using a recurrent neural network (Gated Recurrent Unit; GRU~\cite{GRU}).  An attention mechanism~\cite{Attention} is equipped between the encoder and decoder, and features of each part of the image are selectively sent to the decoder.  The entire system was trained with the MS-COCO dataset~\cite{MS-COCO}.  We followed TensorFlow's tutorial implementation~\footnote{Image captioning with visual attention \url{https://www.tensorflow.org/tutorials/text/image_captioning}\\
The architecture of the decoder is Embed(256)--GRU(512)--FC(512)--FC.}.

\noindent
{\bf BLIP}: This is a foundation model that has learned a large number of images and amount of text and is applicable to a wide range of Vision and Language tasks~\cite{BLIP}.  When used as an image captioner, it takes the form of an encoder-decoder configuration based on the Transformer model.  Users can easily run the sample code (demo.ipynb) on GitHub~\footnote{\url{https://github.com/salesforce/BLIP}} to obtain captions for their own images.  BLIP is a relatively advanced, large-scale model that is capable of producing quite accurate captions.

\noindent
{\bf CLIP Interrogator}: Given an image, this model infers prompts for such AI image generators as Stable Diffusion and Midjourney so as to generate similar images.
Since the text generated by CLIP Interrogator is not meant to be read by humans, it may not be considered an image captioner in the strict sense, but we tested it as a model that can generate richer text than can BLIP.

Although the technical specifications of CLIP Interrogator have not been published as a paper and there is no literature that can be referred to, it may be presumed from the code~\footnote{\url{https://github.com/pharmapsychotic/clip-interrogator}} and its operation that it first generates a base caption using BLIP and then selects and adds phrases that match the target image from a predefined set of phrases called {\it Flavors}.  CLIP image/text encoders~\cite{CLIP} are used to measure the degree of matching between a target image and the phrases in {\it Flavors}.  {\it Flavors} contains approximately 100,000 words and phrases, including those referring to objects and entities (e.g., motorcycle, building, young woman), image styles (e.g., photo-realistic), and artist names (e.g., greg rutkowski).  We used the code released by the developer (clip\_interrogator.ipynb, version 2.2).
\vspace{3mm}

It should be noted that the default settings are different for the standalone BLIP and the BLIP running within the CLIP Interrogator, and that the latter emphasizes the quality of captions by changing search parameters (e.g., num\_beams).  In our experiment, in order to match their operations, the parameters of the standalone BLIP were matched with those of the CLIP Interrogator.

\subsection{System Fusion}
As previously indicated in Figure~\ref{fig:system architecture}, we fuse the outputs of an image-based classifier with those of a text-based classifier to improve classification accuracy.  Fusion methods include feature-level fusion (early fusion), which inputs the hidden layer states of each classifier into another neural-network classifier, and score-level fusion (late fusion), which averages the classification results of each classifier.  Here we choose the latter for simplicity.  That is, if the number of classes is $C$ and the output of the image/text-based classifiers normalized by the softmax function are
$\bm{y}^{\left(I\right)}=\left(y_1^{\left(I\right)},\cdots,y_C^{\left(I\right)}\right)$ and
$\bm{y}^{\left(T\right)}=\left(y_1^{\left(T\right)},\cdots,y_C^{\left(T\right)}\right)$,
respectively, then the classification result obtained with score-level fusion may be calculated as
$\bm{y}=\left(1-w\right)\bm{y}^{\left(I\right)}+w\bm{y}^{\left(T\right)}$, where $0 \le w \le 1$ is the weight coefficient for the text-based classifier.

%------------------------------------------------------------------------- 
\begin{table}[b!]
\caption{Number of images and classes contained in two tasks defined in the CrisisNLP dataset.}
\label{tab:dataset}
\begin{center}
\begin{tabular}{|p{9mm}||p{30mm}|p{30mm}|} \hline
Task    & Disaster types & Damage severity \\ \hline\hline
Train   & 12,724 & 26,898 \\
Dev     &  1,574 &  2,898 \\
Test    &  3,213 &  5,100 \\ \hline
Num classes &
7 (earthquake, fire, flood, hurricane, landslide, other disaster, not disaster) & 
3 (severe damage, mild damage, little or none) \\ \hline
\end{tabular}
\end{center}
\end{table}

\section{Experiments}
\label{sec:experiments}
\begin{figure}[tbp]
\begin{center}
\includegraphics[scale=0.45]{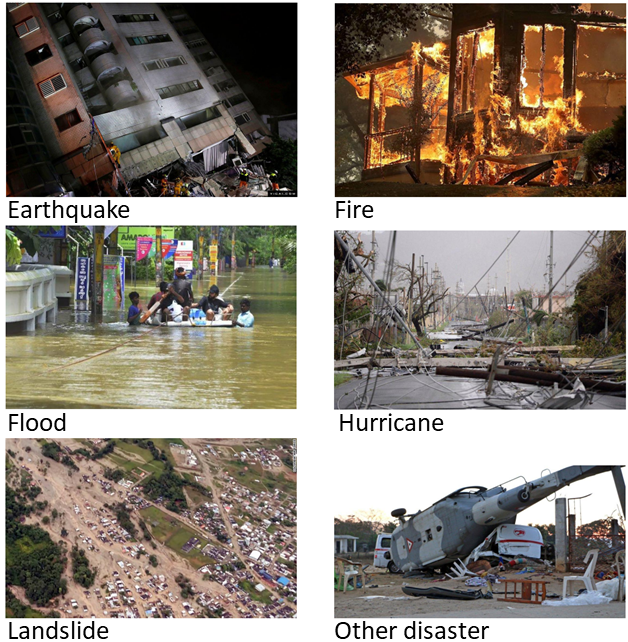}
\end{center}
%\vspace{-5mm}
\caption{Example images for different disaster types included in the CrisisNLP dataset (cited from \cite{CrisisNLP}). There is another type referred to as "not disaster," which is not shown here.}
\label{fig:CrisisNLP}
\end{figure}

We used the CrisisNLP dataset~\cite{CrisisNLP}, which is a collection of natural disaster images shared on such social media as Twitter~\footnote{\url{https://crisisnlp.qcri.org/crisis-image-datasets-asonam20}}.  The dataset provides four image classification tasks, for each of which training (Train), development (Dev), and test (Test) data partitions are defined.  Among them, we focus on two tasks: 
1) ``Disaster types,'' for predicting types of disasters, such as earthquake, flood, etc.;
2)  ``Damage severity,'' for predicting the degree of damage caused by a disaster in terms of three stages. (See Table~\ref{tab:dataset} and Figure~\ref{fig:CrisisNLP})

\begin{table}[tbp]
\caption{Accuracies (\%) of image-based classifiers: MobileNetV2 (MobNetV2), EfficientNet (EffNet) and text-based classifiers: BERT combined with InceptionV3+RNN (IV3+RNN), BLIP (BLIP), CLIP-I (CLIP Interrogator).}
\label{tab:single modal systems}
\begin{center}
\begin{tabular}{|l|cc|} \hline
System & Disaster types & Damage severity \\ \hline\hline
{\em Image-based} & & \\
\hspace{2mm} MobNetV2  & 69.79 & 73.37 \\
\hspace{2mm} EffNet (B0) & {\bf 76.48} & {\bf 76.84} \\
\hspace{2mm} EffNet (B1) & 75.05 & 76.71 \\
\hspace{2mm} EffNet (B2) & 76.33 & 76.31 \\ \hline
{\em Text-based}  & & \\
\hspace{2mm} IV3+RNN & 42.40 & 52.36 \\
\hspace{2mm} BLIP    & 71.14 & 72.17 \\
\hspace{2mm} CLIP-I  & {\bf 85.28} & {\bf 78.67} \\ \hline
\end{tabular}
\end{center}
\end{table}

We first show the classification accuracies of single-modal systems using only an image-based classifier or a text-based classifier (Table~\ref{tab:single modal systems}).  To reduce the randomness of model parameter initialization in fine-tuning, each of those accuracies is averaged over five trials.

\begin{figure}[t]
\begin{center}
\begin{tabular}{|p{15mm}|p{56mm}|} \hline
Input & 
\raisebox{-0.97\height}{\includegraphics{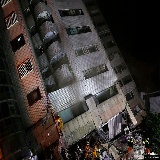}} \\ \hline
MobNetV2 & $\rightarrow$ {\bf not disaster} \\ \hline
EffNet B0 & $\rightarrow$ {\bf earthquake} \\ \hline
IV3+RNN & a restaurant that are shining that is lined up with lots of people. \\
& $\rightarrow$ {\bf not disaster} \\ \hline
BLIP & a group of people standing on top of a building \\
& $\rightarrow$ {\bf not disaster} \\ \hline
CLIP-I & a group of people standing on top of a building, collapsed building, buildings collapsed, collapsed buildings, videogame still, burning building, building destroyed, background of resident evil game, 19xx :2 akira movie style : 8, photo”, big impact hit on the building, damaged buildings, earthquake, unreal engine. film still, test, building on fire \\
& $\rightarrow$ {\bf earthquake} \\ \hline
\end{tabular}
\end{center}
\caption{Example results with an image to be classified as ``earthquake.''}
\label{fig:Example result}
\end{figure}

Regarding the image-based classifiers, EfficientNet models of three sizes (B0, B1, B2) outperformed MobileNetV2 while no significant difference was observed among the three.  Comparing the text-based classifier with three different image captioners, we can first see that the most basic image captioner, InceptionV3+RNN, falls far short of the classification accuracy of standard image-based classifiers.  A look at the captions generated by InceptionV3+RNN reveals that most of them are seemingly irrelevant with respect to the images, and it seems difficult to predict either the type of disaster or the degree of damage from these captions (as indicated later in Figure \ref{fig:Example result}).  By way of contrast, BLIP did generate good captions for many images.  The caption in previously shown in Figure~\ref{fig:system architecture} is one actually generated by BLIP, and it notes such important elements in the image as "people" and "rubble."  The text-based classifier with BLIP consequently achieved much better accuracy, roughly comparable to that of standard image-based classifiers.  The accuracy of the text-based classifier using CLIP Interrogator (CLIP-I) is even higher, and results significantly exceed those of standard image-based classifiers.  These results suggest that foundation models of Vision and Language trained on a large amount of image/text data can be an effective feature extractor for image classification.

Figure~\ref{fig:Example result} shows an example of image classification results for an image that should be classified as ``earthquake.''  As previously noted, InceptionV3+RNN (IV3+RNN), which was the most basic image captioner, produced a description that was far from the actual content of the image.  BLIP's descriptions were generally much more accurate.  The CLIP Interrogator (CLIP-I) behaved quite differently from these two.  After beginning with a common sentence from BLIP, it continued the explanation with a large number of phrases selected by CLIP.  Although those phrases make no sense as sentences, we can observe some that reflect the true class, such as ``earthquake'' and ``collapsed building.''  On the other hand, there are also such completely irrelevant phrases as ``akira movie style'' and ''videogame still;'' they would, however, be good clues for Stable Diffusion to use to generate images. %このような無関係が語句が悪影響を及ぼしているのか, あるいは逆にデータ拡張のような役割を果たしているのか.

\begin{figure}[t]
\begin{flushleft}
(a) Disaster types
\end{flushleft}
\begin{center}
% 563x432
\includegraphics[scale=0.65]{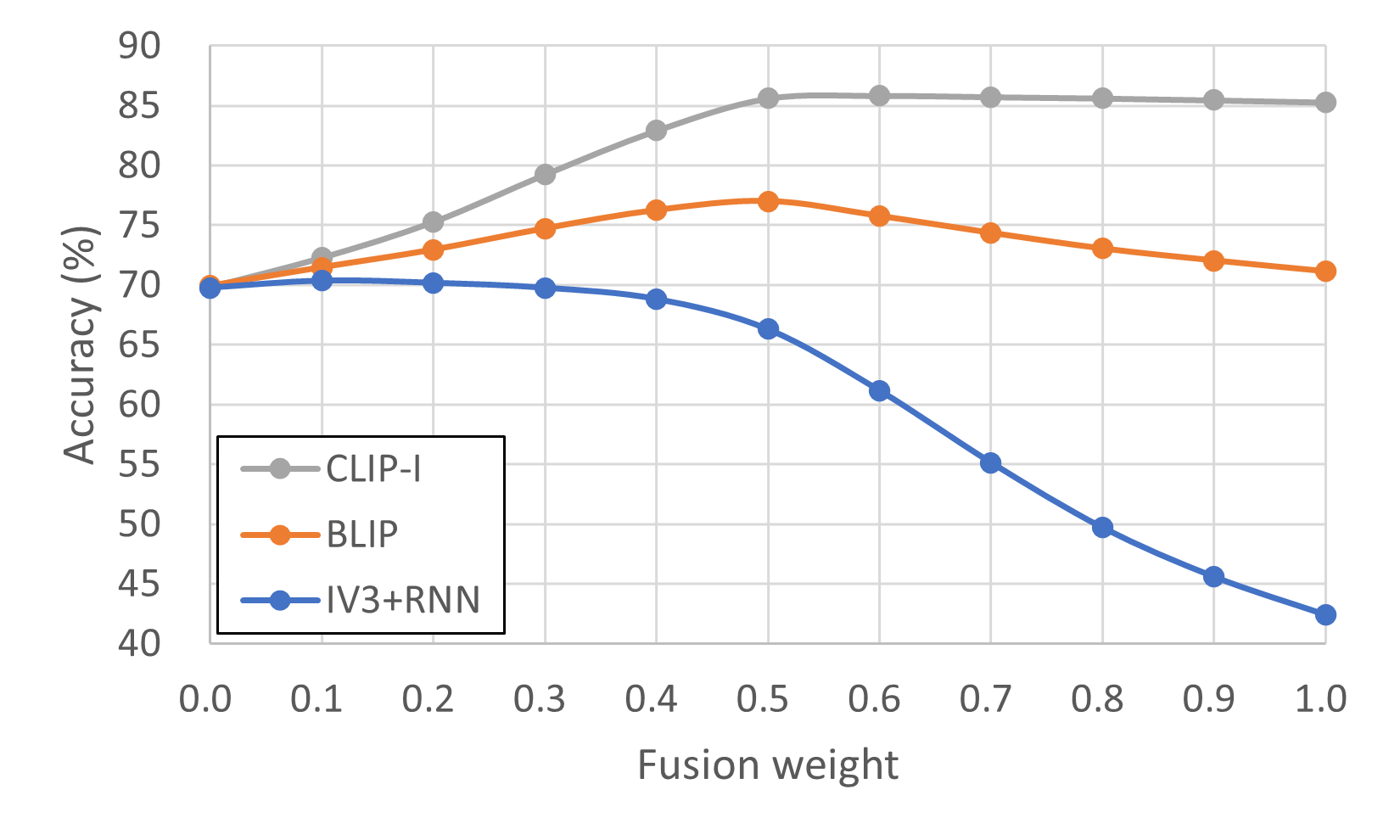}
\end{center}
\begin{flushleft}
(b) Damage severity
\end{flushleft}
\begin{center}
%% 563x438
\includegraphics[scale=0.65]{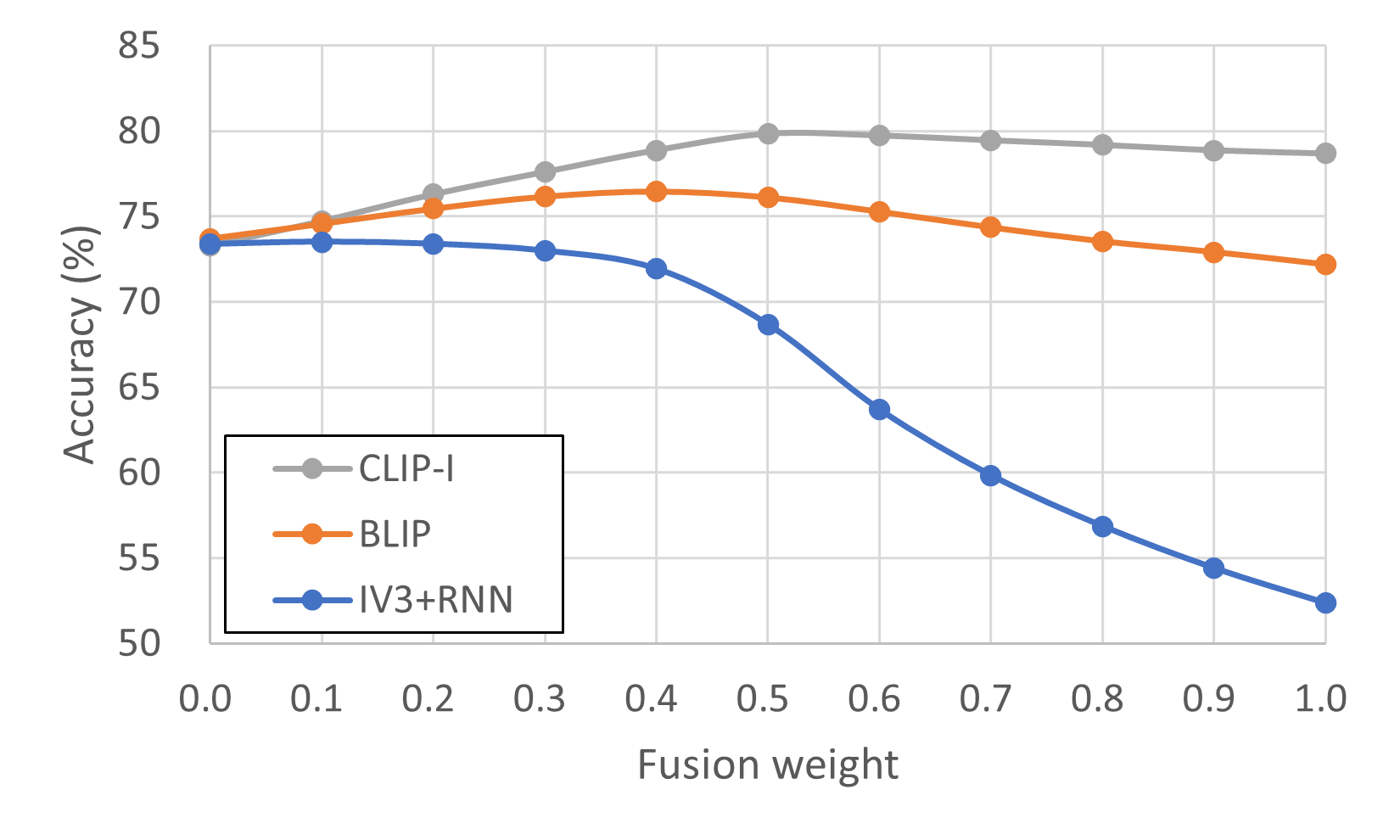}
\end{center}
\caption{Score-level fusion results using MobileNetV2 as the image-based classifier: Horizontal axis stands for the fusion weight $w$ for the text-based classifier.  $w=0$ and $w=1$ correspond to image-based and text-based single-modal systems, respectively.}
\label{fig:Fusion_MobNetV2}
\end{figure}

Figure~\ref{fig:Fusion_MobNetV2} shows the results of score-level fusion that averaged the output of the image/text-based classifiers with weight $w$, where MobileNetV2 was used for the image-based classifier.  Both of the two tasks (Disaster types, Damage severity) show similar trends, i.e., when using sufficiently good models, such as BLIP and CLIP Interrogator (CLIP-I), for image captioning, the classification accuracy can be improved by appropriately choosing the weight $w$.  It seems that image captioning models extract features different from CNN models; in other words, they look at images from a different perspective than do CNNs.  Unfortunately, when using InceptionV3+RNN for image captioning, the effect of fusion was nealy unseen because the difference in performance between the two modalities was too large.

\begin{figure}[t]
\begin{flushleft}
(a) Disaster types
\end{flushleft}
\begin{center}
% 563x432
\includegraphics[scale=0.65]{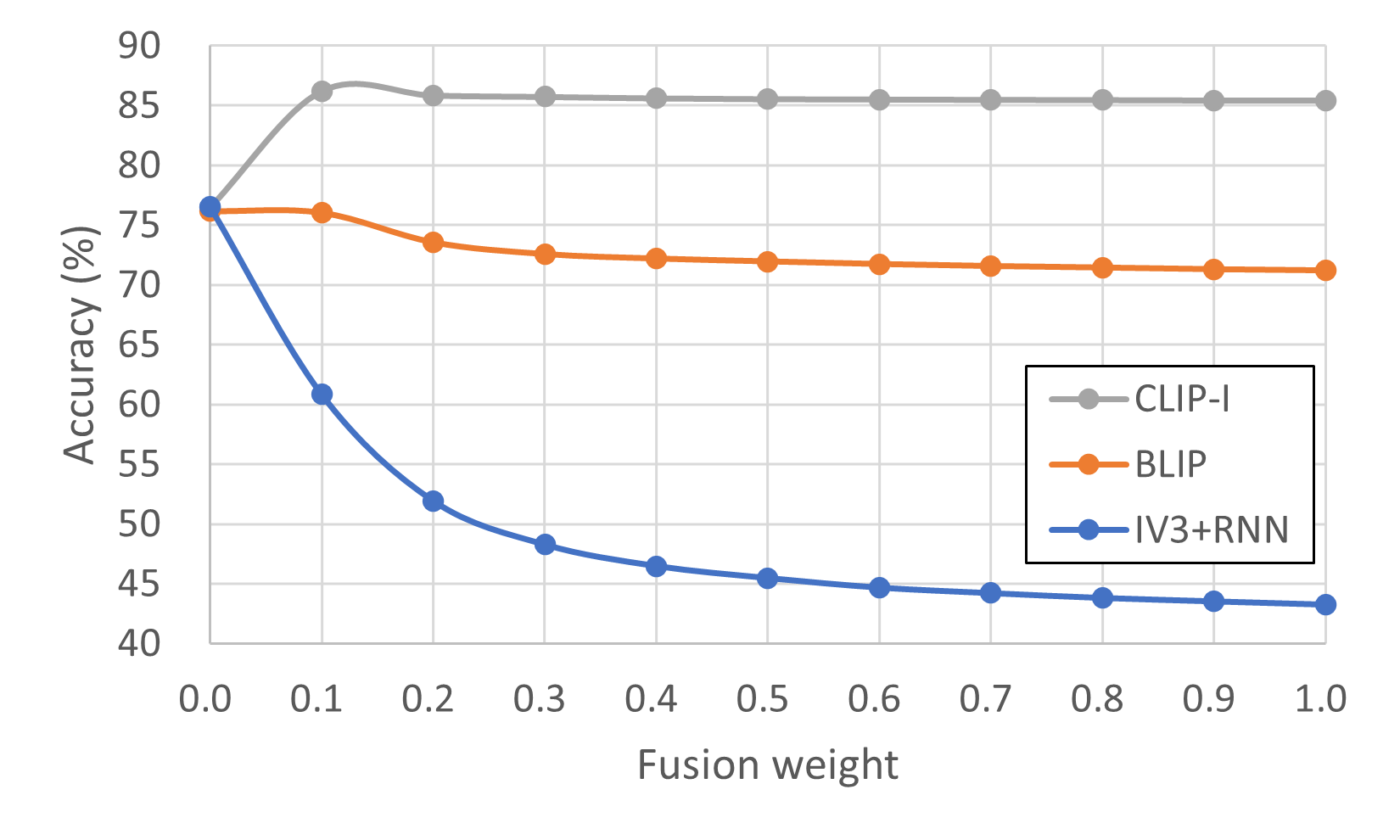}
\end{center}
\begin{flushleft}
(b) Damage severity
\end{flushleft}
\begin{center}
%% 563x438
\includegraphics[scale=0.65]{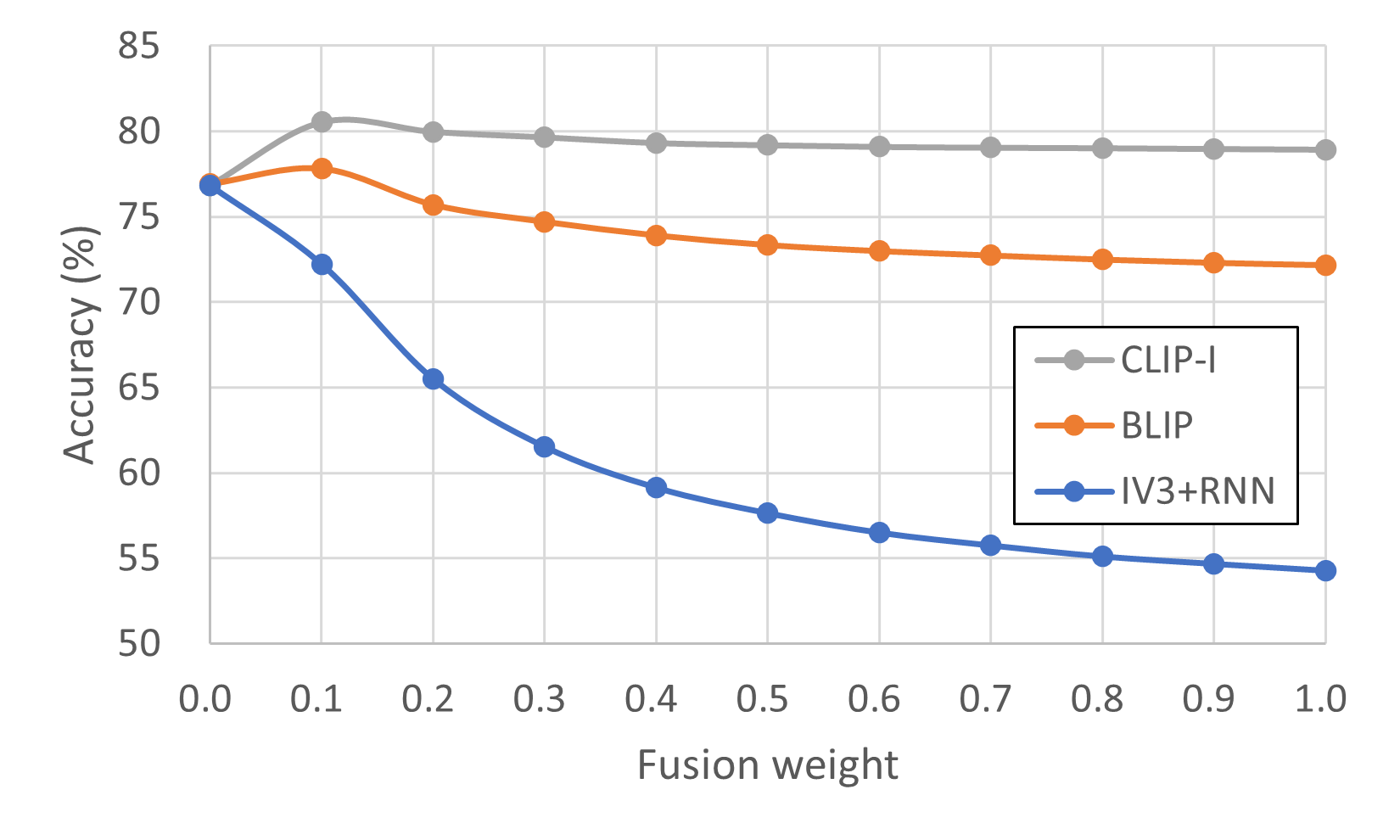}
\end{center}
\caption{Score-level fusion results using EfficientNet B0 (rather than MobileNetV2 in Figure~\ref{fig:Fusion_MobNetV2}) as the image-based classifier.}
\label{fig:Fusion_EffNet}
\end{figure}

When we fused the text-based classifier with another image-based classifier, EfficientNet (B0), we could still see a synergistic effect, as indicated in Figure~\ref{fig:Fusion_EffNet}.  We should note that, as the classification accuracy of EfficientNet is higher than that of the text-based classifier using BLIP (unlike MobileNetV2), the synergistic effect is not so large as in the case of MobileNetV2.

%------------------------------------------------------------------------- 
\section{Summary}
\label{sec:conclusion}
Starting from the question of what kind of information image captioners can extract from images, we performed image classification using only linguistic information contained in captions, and have shown that it is possible to achieve classification accuracy that surpasses that of standard image-based classifiers using CNNs.  Further, we have confirmed that synergistic effects can be obtained by fusion with those image-based classifiers.  It can be said that image captioners based on large-scale foundation models are effective feature extractors for image classification.  Image captioning is a rapidly evolving area, and we need to keep up with new technologies that are being released one after another~\cite{BLIP-2}~\footnote{\url{https://docs.midjourney.com/docs/describe}}.

In this study, we experimented with a simple system configuration, in which very basic parts had been combined.  In future work, we intend to introduce more advanced methods, such as feature-level fusion and knowledge distillation~\cite{Srinivasan}, in order to search for better answers to the original question.  It should be further noted that the behavior of image captioners would seem to depend heavily on the images to be captioned.  While scenes of disasters are relatively easy to describe, the characteristics of a human face would not be so easy.  Studies on other datasets are to be included in our future work.

%------------------------------------------------------------------------- 
\section*{Acknowledgment}
This work was partially supported by JSPS KAKEN Grant Number 21K11967.

%------------------------------------------------------------------------ 
%\nocite{ex1,ex2}
%\newpage
\bibliographystyle{latex8}
\bibliography{latex8}

\end{document}